\documentclass[10pt,twocolumn,letterpaper]{article}

\usepackage[pagenumbers]{cvpr} %

\definecolor{cvprblue}{rgb}{0.21,0.49,0.74}
\newcommand{\R}{\mathbb{R}}
\usepackage[pagebackref,breaklinks,colorlinks,allcolors=cvprblue]{hyperref}
\usepackage{makecell}
\usepackage{colortbl}
\usepackage{enumitem}
\newcommand{\best}{\cellcolor{tablered}}
\newcommand{\sbest}{\cellcolor{orange}}
\newcommand{\tbest}{\cellcolor{yellow}}

\definecolor{yellow}{rgb}{1, 1, 0.7}
\definecolor{orange}{rgb}{1, 0.85, 0.7}
\definecolor{tablered}{rgb}{1, 0.7, 0.7}

\def\paperID{15087} %
\def\confName{CVPR}
\def\confYear{2025}

\begin{document}

\title{GaussianVideo: Efficient Video Representation via Hierarchical Gaussian Splatting}

\author{
    Andrew Bond\textsuperscript{1,*} \quad Jui-Hsien Wang\textsuperscript{2} \quad Long Mai\textsuperscript{2} \quad Erkut Erdem\textsuperscript{3} \quad Aykut Erdem\textsuperscript{1} \vspace{1mm} \\
    \textsuperscript{1}Koç University \quad \textsuperscript{2}Adobe Research \quad \textsuperscript{3}Hacettepe University\vspace{2mm}\\
    \textbf{\url{https://cyberiada.github.io/GaussianVideo/}}
\vspace{-0.4cm}
}
\makeatletter
 \g@addto@macro\@maketitle{
     \centering
   \includegraphics[width=0.99\textwidth]{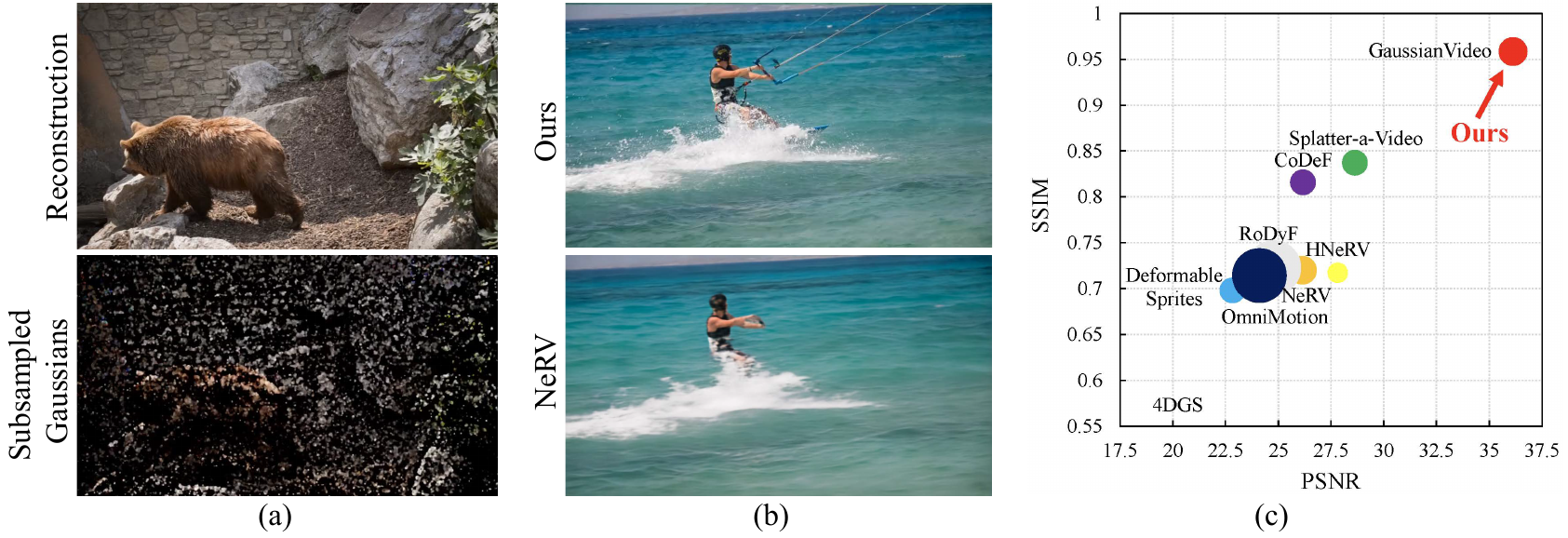}
   \vspace{-4mm}
   \captionof{figure}{\textbf{We present GaussianVideo -- a new Gaussian Splatting framework for representing videos} that effectively models in-the-wild videos, while maintaining training efficiency and capturing semantic motions with minimal supervision. (a) We can render the 960$\times$540 video at 93~FPS on an NVIDIA A40 GPU. (b) Our reconstruction PSNR on this video achieves 44.21 compared to NeRV~\cite{NeRV} at 29.36, representing a 50.6\% improvement. (c) On the DAVIS~\cite{davis} dataset, our approach balances reconstruction quality with training time (dot size in log scale).}
   \label{fig:teaser}\vspace{5pt}
 }
 \makeatother
 \maketitle

 {\let\thefootnote\relax\footnotetext{{*} Done during an internship at Adobe Research.}}

 \begin{abstract}
Efficient neural representations for dynamic video scenes are critical for applications ranging from video compression to interactive simulations. Yet, existing methods often face challenges related to high memory usage, lengthy training times, and temporal consistency. To address these issues, we introduce a novel neural video representation that combines 3D Gaussian splatting with continuous camera motion modeling. By leveraging Neural ODEs, our approach learns smooth camera trajectories while maintaining an explicit 3D scene representation through Gaussians. Additionally, we introduce a spatiotemporal hierarchical learning strategy, progressively refining spatial and temporal features to enhance reconstruction quality and accelerate convergence. This memory-efficient approach achieves high-quality rendering at impressive speeds. Experimental results show that our hierarchical learning, combined with robust camera motion modeling, captures complex dynamic scenes with strong temporal consistency, achieving state-of-the-art performance across diverse video datasets in both high- and low-motion scenarios. %
\end{abstract}
    
 \section{Introduction}
\label{sec:intro}

Videos are traditionally represented as a sequence of frames, each of which is a 2D grid of discrete pixels with fixed resolutions. %
Recently, it has become feasible to model the continuous nature of videos using implicit representations such as NeRFs~\cite{NeRV}, by modeling the videos as neural networks that output whole frames. It was shown that this is a promising direction in replacing the decade-old frame-based representation, and encoding and decoding speed can both be improved along with reconstruction quality.

On the other hand, 3D Gaussian Splatting (3DGS)~\cite{3dgs} has been gaining popularity in modeling 3D scenes as an alternative implicit representation. Compared to NeRFs, 3DGS-based methods enable faster rendering speed and smaller memory usage, and their continuous and interpretable representations allow applications such as stylization~\cite{G-Style}, 3D editing~\cite{gaussianeditor}, frame interpolation~\cite{splatter-a-video}, novel-view synthesis~\cite{3dgs}, and inverse rendering~\cite{gs-ir}. 
Recently, GaussianImage~\cite{gaussianimage} showed that Gaussian splatting can also be applied to represent images. Leveraging the fixed-viewpoint nature of images, and optimization techniques and codecs designed specifically for Gaussians, they showed that replacing pixels with Gaussians can be feasible and more efficient for rendering images. Furthermore, Splatter-a-Video~\cite{splatter-a-video} explored modeling videos using Gaussians. They showed that by utilizing a myriad of supervisory signals such as optical flow, depth, segmentation maps, and others at training, the obtained GS-based representation can be very versatile for downstream processing while maintaining good reconstruction quality.

\begin{figure}[t!]
    \centering
    \includegraphics[width=\linewidth]{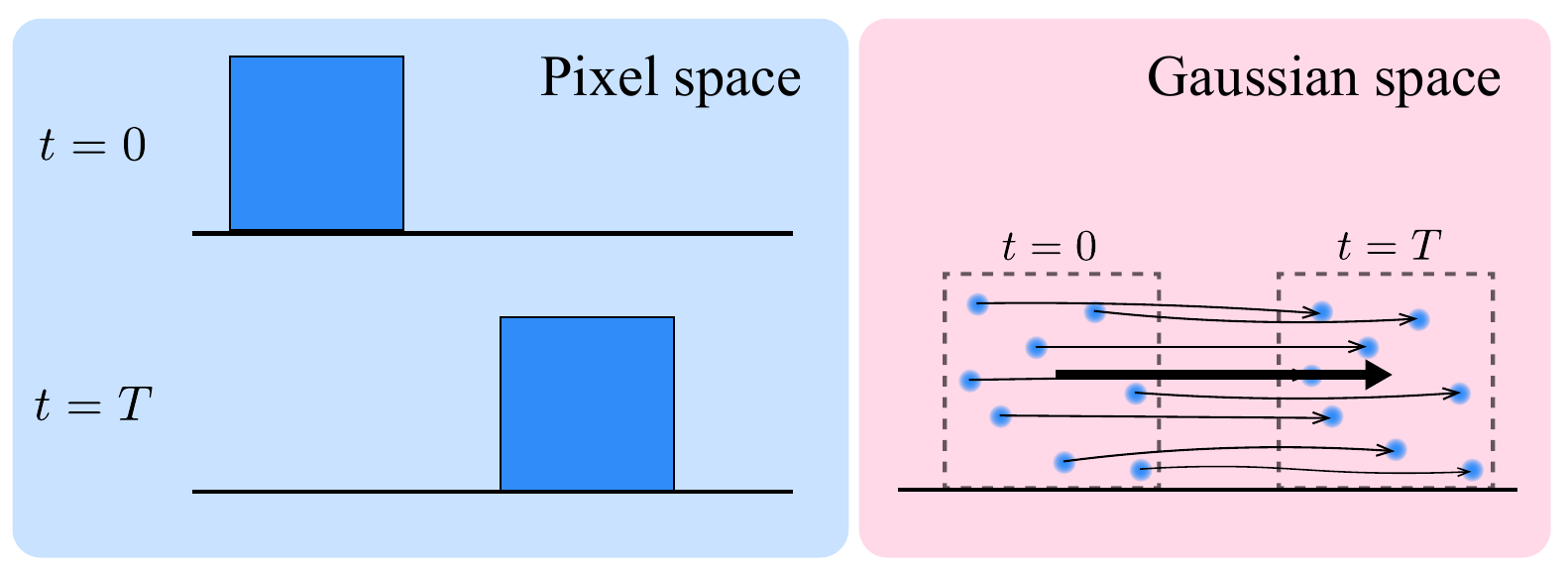}
    \caption{\textbf{A toy example illustrating the semantic and coherent motion of the underlying Gaussians}. Rather than relying on strong supervisory signals like optical flow during training, we carefully design a Gaussian parametrization tailored for video modeling to naturally encourage this behavior, and demonstrate its effectiveness in downstream applications.}
    \label{fig:semantic_illustration}
\end{figure}

In this paper, we introduce a lightweight framework for modeling videos using Gaussians. 
Similar to Splatter-a-Video, our design goals are twofold: (1) to achieve high-quality and efficient video reconstruction, and (2) to support downstream applications such as frame interpolation, spatial resampling, editing, and more. We posit that enabling these applications requires the dynamics of the Gaussians to be highly semantic and coherent. Unlike Splatter-a-Video, we seek an approach that results in this type of semantic movement of the Gaussians as an \emph{emerging} behavior, rather than being forced by the auxiliary supervisory signals (see Fig.~\ref{fig:semantic_illustration} for illustration). This is because these supervisory signals can themselves be expensive to compute and challenging to be accurately extracted; in addition, they might hide the negative effects of the design choices made in the representation itself. For this purpose, we propose several novel techniques to model the videos and regularize the movements of Gaussians. The main technical contributions of the paper are as follows: 
\begin{itemize}
\item{We introduce a B-spline-based motion representation that models smooth motion trajectories of scene elements, ensuring temporal consistency while allowing for local variations in motion (\S\ref{sec:video_representation})}
\item{We propose a novel hierarchical learning strategy that refines spatial and temporal features of the Gaussian representation, leading to improved reconstruction quality and faster convergence compared to existing methods (\S\ref{sec:hierarchical_gaussians})}
\item{We develop an approach to learning continuous camera motion using Neural ODEs, removing the dependency on precomputed camera parameters and enhancing adaptability to different capture setups (\S\ref{sec:model_camera}).}
\item{Our method significantly reduces memory usage and training time, achieving state-of-the-art performance on standard video datasets with lower computational resources compared to previous approaches.}
\end{itemize}

\begin{figure*}[t!]
    \centering
    \includegraphics[width=\linewidth]{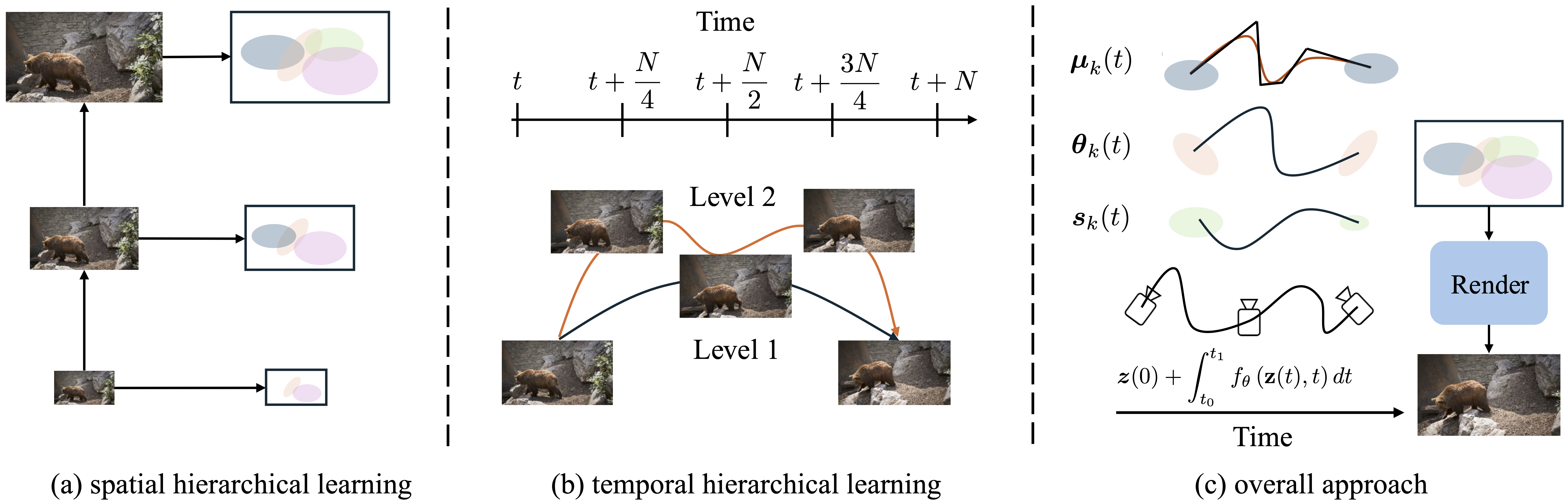}
    \caption{\textbf{Overview of the GaussianVideo approach for neural video representation.} Our method combines 3D Gaussian splatting with continuous camera motion modeling via Neural ODEs to handle dynamic scenes efficiently. The pipeline includes hierarchical learning strategies for both (a) spatial and (b) temporal domains, progressively refining Gaussians to capture fine details and smooth motion.}
    \label{fig:overview}
\end{figure*}

An overview of our approach is illustrated in Fig.~\ref{fig:overview}. Our method surpasses existing approaches in reconstruction quality and temporal consistency, effectively handles videos with varying levels of camera and object motion, and achieves competitive training times and memory efficiency.

 \section{Related Work}
\label{sec:related}

\subsection{Implicit Neural Video Representations}
Recent work have shown that neural networks can be used to represent various types of multimedia signals~\cite{siren}, including images~\cite{inrcompress}, videos~\cite{NeRV}, and audio~\cite{inras}. For videos, in contast to traditional frame-based video representations, these neural video representations provide more compact and flexible modeling. NeRV \cite{NeRV} was the first implicit representation that models an entire video using MLPs by mapping timesteps to frames. HNeRV \cite{HNeRV} extends this approach by employing a hybrid representation with content-adaptive embeddings, alongside a balanced parameter distribution across the network, enhancing efficiency and adaptability. DS-NeRV \cite{DSNeRV} decomposes videos into static and dynamic codes, which are co-optimized during training to capture both persistent and transient video elements. Meanwhile, Splatter-a-Video \cite{splatter-a-video} uses Gaussian splatting to represent videos, but with supervision such as segmentation masks, depth maps, optical flow, and feature maps. 

\subsection{Dynamic Gaussian Splatting}
3D Gaussian Splatting (3DGS)~\cite{3dgs} was originally introduced as a method to learn static 3D scenes. However, significant efforts have since extended this approach to dynamic scenes. Spacetime GS \cite{spacetime} models the positions and rotations with polynomials, represents colors through an MLP, and employs radial basis functions for opacities. 4DGS \cite{4dgs} advances this framework by utilizing a learned deformation network parametrized by an MLP to predict the offsets of position, rotation, and scales over time. Meanwhile, DynMF \cite{dynmf} uses a learned motion basis of distinct motions in a scene, with each motion trajectory parameterized by a small MLP. MotionGS \cite{motiongs} employs optical flow to decouple camera motion from object motion, allowing for more constrained deformation of Gaussians. Recently, constrained approaches to dynamic Gaussian splatting, such as \cite{V3_gs}, have emerged to efficiently learn volumetric videos with a reduced Gaussian count.
In this paper, we focus on adapting the 3DGS formulation to model video data instead of modeling dynamic 3D scenes.

\subsection{Camera Modeling}
The standard approach for obtaining camera information from images is COLMAP \cite{colmap}, a type of Structure-from-Motion (SfM) algorithm. However, recently there have been attempts to replace COLMAP with a deep-learning based approach. Dust3r \cite{dust3r} uses an unconstrained image collection to predict a point cloud as well as intrinsic and extrinsic camera information, but crucially requires at least two images. Alternatively, some works, such as \cite{colmap-free}, try to avoid the use of COLMAP by progressive training which predicts Gaussian transformations between steps. 

 \section{Method}
\label{sec:approach}

\subsection{Background}
\label{sec:background}
\noindent\textbf{Gaussian Splatting.} Gaussian Splatting reconstructs a scene in $d$ dimension by optimizing a set of Gaussians~\cite{3dgs}, each of which is parameterized by position $\mathbf{\mu} \in \R^{d}$, covariance $\mathbf{\Sigma} \in \R^{d\times d}$, colors with spherical harmonics $\textbf{c} \in \R^{(d_c + 1)^{2}}$, and opacity $o \in \R$. $d_c$ is the order of spherical harmonics. In the forward pass, the Gaussians are projected and rasterized onto the observed image planes, and through standard reconstruction loss the parameters can be optimized. The covariance matrix can be further parametrized using a decomposition $\Sigma = (\textbf{R}\textbf{S})(\textbf{R}\textbf{S})^{T}$, where $\textbf{R}$ is a rotation matrix and $\textbf{S}$ is a (diagonal) scaling matrix.

\vspace{1mm}\noindent\textbf{Neural ODEs.}
A Neural ODE \cite{neural_ode} is a continuous-depth model in which the continuous dynamics of the hidden units are parametrized using a differential equation, where the function being differentiated is a neural network. Specifically, given some initial condition $\textbf{z}(0)$, the Neural ODE solves the following equation
\begin{equation}
    \textbf{z}_{dT} = \textbf{z}(0) + \int_{t_{0}}^{t_{1}} f_{\theta}(\textbf{z}(t), t)dt,
\end{equation}
where $f_\theta$ is the network parametrization. The initial condition can be learned, or predicted using another network.

\subsection{Adding Dynamics to the Gaussians}
\label{sec:video_representation}

Videos contain a lot of redundant information. For example, frames in a video of someone walking down the street will share many similarities, such as the color of the shirt or the shades in the sky. To model this temporal aspect, we augment the parameters of the Gaussians with temporal basis functions. Below, we discuss various choices and explain our design choices.

\begin{figure}[t!]
    \centering
    \includegraphics[width=1.0\linewidth]{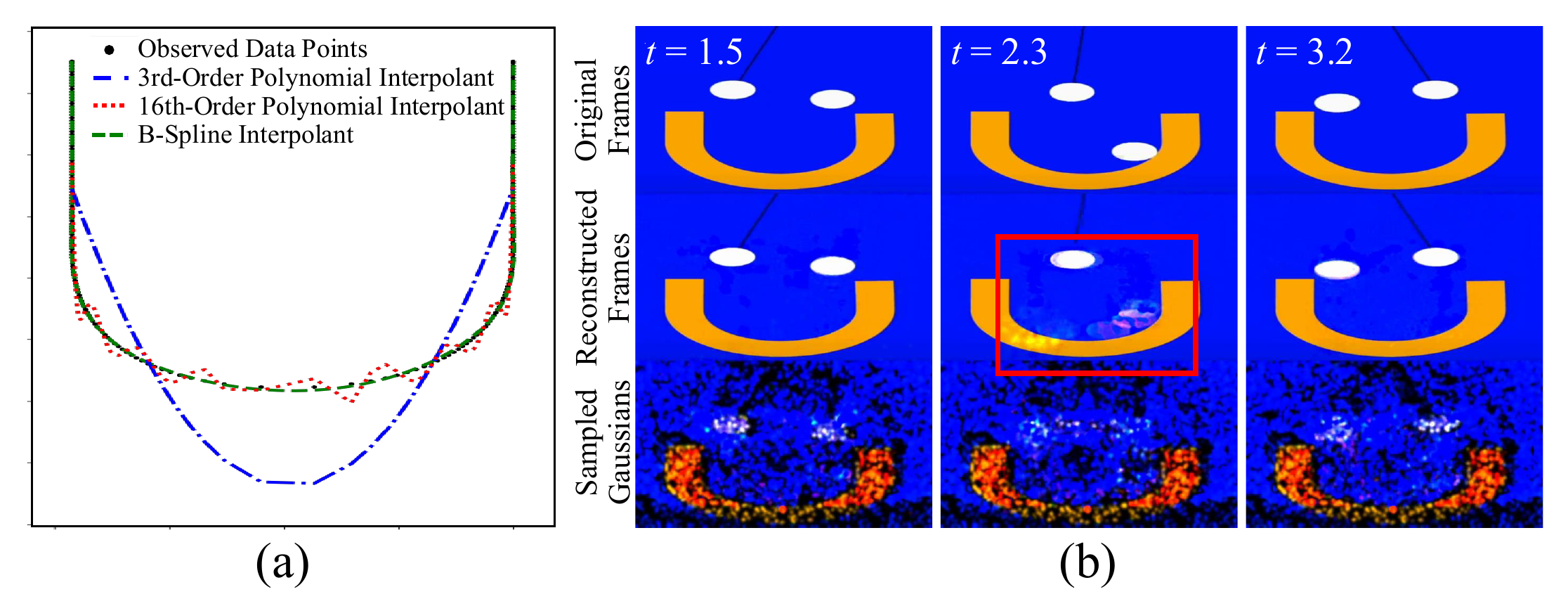}
    \caption{\textbf{Polynomial basis functions, widely used for temporal modeling, can introduce reconstruction errors and instability}, as shown in this example. (a) 3rd-order polynomial fails to fit the U-shape trajectory, while the commonly used 16th-order polynomial fits better, but is unstable and sensitive to small perturbations (see the bottom part of the ``U"). In contrast, our cubic B-spline formulation (green dashed lines) fits the shape is unstable and sensitive to small perturbations due to its flexibility. (b) When using a low-order polynomial for motion fitting, the white ball following the ``U" disappears entirely in the middle frame and reappears later, highlighting the instability of single-polynomial modeling.\vspace{-0.4cm}}
    \label{fig:u}
\end{figure}

We found that the conventional polynomial basis such as those used in~\cite{spacetime,splatter-a-video,lin2023gaussian} to model positional changes can cause overfitting and reconstruction errors for even simple trajectories such as a U-shape (see Fig.~\ref{fig:u}). Unsurprisingly, this is because a single polynomial at low order does not approximate complex trajectories well, especially those from a video of long duration or high dynamics; at high order, polynomials are known to be unstable and sensitive to noise. Fortunately, this problem is well studied in computer graphics and approximation theory~\cite{marschner2016graphics}. We show that a simple adoption of basic cubic B-splines can greatly alleviate this issue (see Fig.~\ref{fig:u}). Specifically, we model the positions of the $n$-th Gaussians as a time-dependent function
\begin{equation}
    \mu_{n}(t) = \sum_{i=0}^{N} N_{i, p}(t)\textbf{P}_{n, i},
\end{equation}
where $\textbf{P}_{n, i} \in \R^{d}$ is the i-th control point, and $N_{i, p}$ are the $p$-th degree basis functions. We use $p=3$ for a common cubic spline. $N$ is the number of control points. We use clamped B-splines and thus the first $p+1$ knots are fixed at $0$, and the last $p+1$ knots are fixed at $1$; the rest of the knots are evenly spaced.

In contrast, for other degrees of freedom, such as covariance (through rotation and scaling matrices), color, and opacity, we intentionally use less expressive models. This is because (1) if opacity and scaling are allowed to change dramatically, the reconstruction loss tends to push the Gaussians to become insignificant. For example, if the opacity of a Gaussian is allowed to vary over time, we found that on average $80-85\%$ of the Gaussians will have low opacity ($o < 0.5$). Similarly, if scaling is allowed to change dramatically, Gaussians tend to shrink in size and become insignificant. Both cause inefficient utilization of the available degrees of freedom and will negatively impact convergence and training time. We therefore do not alxlow opacity to change over time, resulting in much fewer Gaussians having low opacity ($<5\%$); for covariances, we use a regular 3rd-order polynomial to model the scaling and rotation, similar to ~\citet{spacetime}. (2) A popular choice for modeling the color of the Gaussians is through a learned MLP~\cite{spacetime}; however, we found that even a weak MLP encourages the Gaussians to change colors rather than move around to follow entities in the video. We therefore do not allow the color to change over time either. More analysis of these design choices is provided in the supplemental material.%

In summary, the dynamics of the $n$-th Gaussian is parametrized by
\begin{align}
    \mathcal{\tilde{G}}_n(t) = \mathcal{G}_n\left(\underbrace{\mathbf{\mu}_n(t)}_{\text{B-splines}}, 
    \underbrace{\mathbf{\Sigma}_n(t)}_{\text{3rd-order polynomials}}, 
    \underbrace{\textbf{c}_n}_{\text{constant}}, 
    \underbrace{o_n}_{\text{constant}}\right)
\end{align}
to properly capture the motions in the videos yet maintain modeling efficiency and semantics of the representation.

\subsection{Hierarchical Representations of the Gaussians}
\label{sec:hierarchical_gaussians}

Inspired by the Multigrid methods~\cite{Hackbusch1985multigrid} in solving differential equations and other hierarchical representations in the Gaussian Splatting literature~\cite{hamdi2024ges,lu2024scaffoldgs}, we also propose to decompose the optimization into discrete stages, each targeting different levels of details. By building the representation progressively at several different spatial and temporal scales, from coarse to fine, we found that it results in more efficient learning performance.

\vspace{1mm}\noindent\textbf{{Spatial Hierarchical Learning.}} We first introduce our spatial hierarchical learning approach using the Gaussian pyramid. Recall that a Gaussian pyramid contains several levels of the same image, each with a progressively downsampled version of itself using Gaussian kernel~\cite{burt1983pyramid}. This naturally fit into a Gaussian splatting framework because, intuitively, a point at a lower level of the pyramid can be represented by a single Gaussian perfectly at a higher level due to the kernel choice. And when the higher levels get reprojected back to lower levels (higher resolutions), the only thing that needs changing to continue to fit is the single scalar scale of this Gaussian. Using this rationale, our spatial hierarchy starts the training with the highest level of the pyramid (or coarsest spatial details), trains it to convergence. We then move down one level, introduce more Gaussians, and again train it to convergence. We repeat this process until all $N_p$ levels in the pyramid are trained.

\vspace{1mm}\noindent\textbf{{Temporal Hierarchical Learning.}}
To promote smooth motion that is useful for downstream applications such as frame interpolation (\S\ref{sec:applications}), we similarly adopt a progressive frame sampling approach based on B-splines. We start by training on every N-rh frame, then progressively increase the temporal resolution, optionally refining the B-spline knots to introduce additional flexibility without disrupting the learned Gaussian movements.

\vspace{1mm}\noindent\textbf{{Spatio-Temporal Hierarchical Learning.}} By combining these spatial and temporal hierarchies, we can construct a Gaussian video model that captures both global structure and local details, while also exhibiting smooth, natural motion over time. We validate these claims in \S\ref{sec:results}.

\subsection{Modeling Camera Motion}
\label{sec:model_camera}
Our approach described so far performs well on videos without camera motion but is sensitive to even slight camera shifts. 
Existing approaches typically rely on Structure-from-Motion (SfM) algorithms such as COLMAP \cite{colmap} to estimate camera parameters~\cite{splatter-a-video, 3dgs}. However, this can be unreliable for single-camera videos, as the limited perspectives make it difficult to accurately reconstruct the camera motion. Furthermore, COLMAP is a slow process, making it a bottleneck for training efficiency. 

To address these issues, we propose a simple yet effective approach to learning camera parameters directly within our video reconstruction pipeline. Since the forward pass of our framework already encompasses the entire rendering process, the key insight is that both intrinsic and extrinsic camera parameters can be explicitly learned and subsequently used to render the Gaussians. Specifically, we model the intrinsic parameters (i.e., focal length and principal points) with constants held throughout the video; for extrinsic parameters (i.e., rotation and translation of the camera), we model them with neural ODEs. Details for the camera modeling are given in the supplementary material.

Further optimization can be done to disable camera modeling for videos without significant camera motion, such as those of the DAVIS dataset (see Table~\ref{tab:comparisons}), to learn more efficiently. However, for simplicity, all our experiments always utilize the full camera modeling pipeline. We implemented the camera modeling as custom CUDA kernels for efficient forward and backward passes; this tight integration between camera parameters and rendering pipeline allows for optimized training performance by minimizing memory transfer between devices.

 \vspace{-2mm}\section{Results}
\label{sec:results}
\subsection{Experimental Setup}
\noindent\textbf{Datasets.} We validate our approach experimentally on two datasets: DL3DV \cite{dl3dv} and DAVIS Tap-Vid benchmark \cite{davis}. The DL3DV dataset enables testing our approach on high-camera-motion scenes, while the DAVIS dataset 
allows us to assess performance on lower-motion scenes with a focus on fine detail. Furthermore, the DAVIS dataset allows for a comparison with Splatter-A-Video \cite{splatter-a-video}. In particular, we use the 1K split of the DL3DV dataset at 540p resolution, in which the videos are 30 or 60 FPS, and the videos last for between 5-10 seconds on average. Furthermore, while the DL3DV dataset does come with provided camera information, we do not use them in order to ensure our model is able to learn the camera dynamics by itself. To fairly compare against Splatter-a-Video, we use the DAVIS dataset at 480p resolution, with videos lasting for a few seconds at 30 FPS.

\begin{table*}[!t]
    \centering
    \begin{tabular}{lcccc}
    \toprule
         \textbf{Model} & \textbf{PSNR} $\uparrow$ & \textbf{SSIM} $\uparrow$ & \textbf{LPIPS} $\downarrow$ & \textbf{Training Time}\\
        \midrule
         \multicolumn{5}{c}{\textbf{DL3DV Dataset}} \\
         \midrule
         {GaussianVideo} (Ours) & \best{43.21} & \best{0.99} & \best{0.013} & $\sim$ 45 mins\\
         {GaussianImage} \cite{gaussianimage} & \sbest{38.68} & \sbest{0.95} & \sbest{0.097} & $\sim$ 15 mins\\
         {NeRV} \cite{NeRV} & 28.50 & \tbest{0.87} & 0.184 & $\sim$ 45 mins\\
         {HNeRV} \cite{HNeRV} & \tbest{30.32} & 0.86 & 0.183 & $\sim$ 15 mins \\
         {3DGS} \cite{3dgs} & 29.82 & 0.92 & \tbest{0.120} & 2.1 hr\\
        \midrule
        \multicolumn{5}{c}{\textbf{DAVIS Dataset}} \\
        \midrule
        {GaussianVideo} (Ours) & \best{37.38} & \best{0.96} & \best{0.021} & $\sim$ 45 mins\\ 
        {GaussianImage} \cite{gaussianimage} & \sbest{36.25} & \sbest{0.94} & \sbest{0.045} & $\sim$ 15 mins\\
        {Splatter-a-Video} \cite{splatter-a-video}& \tbest{28.63} & \tbest{0.84} & \tbest{0.228} & $\sim$ 30 mins\\ 
        {NeRV} \cite{NeRV} & 26.15 & 0.72 & 0.312 & $\sim$ 45 mins\\
        {HNeRV} \cite{HNeRV}& 27.82 & 0.72 & 0.252 & $\sim$ 15 mins\\
        {4DGS} \cite{4dgs} & 18.12 & 0.57 & 0.394 & $\sim$ 40 mins\\ 
        {RoDyF} \cite{rodyf} & 24.79 & 0.72 & 0.394 & $>$ 24 hrs\\ 
        {Deformable Sprites} \cite{deformable_sprites} & 22.83 & 0.70 & 0.301 & $\sim$ 30 mins\\ 
        {OmniMotion} \cite{omnimotion} & 24.11 & 0.72 & 0.371 & $>$ 24 hrs\\ 
        {CoDeF} \cite{codef} & 26.17 & 0.82 & 0.290 & $\sim$ 30 mins\\ 
        \bottomrule
    \end{tabular}%
    \caption{\textbf{Comparison of our approach with various methods on the DL3DV \cite{dl3dv} and DAVIS \cite{davis} datasets.} Our method achieves the best performance across all metrics among video representation learning models. Most results for the DAVIS dataset are obtained from the Splatter-A-Video work, which doesn't provide parameter counts. The \colorbox{tablered}{best}, the \colorbox{orange}{second best}, and the \colorbox{yellow}{third best} results are highlighted.}
    \label{tab:comparisons}
\end{table*}

\vspace{1mm}\noindent\textbf{Evaluation Metrics.}
Following existing work, we employ the peak signal-to-noise ratio (PSNR), structural similarity index (SSIM)~\cite{wang2004image} metrics, and learned perceptual image patch similarity (LPIPS)~\cite{zhang2018unreasonable} perceptual image quality metric to assess the video reconstruction performance of the approaches. 

\vspace{1mm}\noindent\textbf{Implementation Details.}
We conduct all experiments on both datasets with a consistent set of hyperparameters and experimental setup. Specifically, we use 400K Gaussians, trained for 50K timesteps, with a learning rate of 0.01 that decays exponentially with a factor of $\gamma = 0.99995$. Following \cite{gaussianimage}, we employ the Adan \cite{adan} optimizer. Gaussians are initialized randomly within a radius of three times the camera width/height, ensuring Gaussians are present regardless of the learned camera motion trajectory.

The results reported here are based on the standard L2 reconstruction loss. Although additional regularization objectives can be used for specific downstream tasks, our primary focus is on optimizing reconstruction quality.

We apply the spatial and temporal hierarchical learning procedure once, after $15$K training steps. At this point, adding $100$K new Gaussians to capture fine details from upsampling. Additionally, every $5$K steps, we identify all Gaussians that do not appear on any frames of the video, and \emph{warp} them to the first frame by adjusting the splines of their means. During this warping, we identify the regions of high error and move them to these regions. This procedure is stopped after 30K steps, as camera motion changes minimally beyond this point.

Finally, while it is common when using spherical harmonics to use a progressive learning scheme that first learns the 0th-order coefficients and gradually increases the order over time, we found that this performs worse than learning all the coefficients at once. This is due to the camera position changing frequently during the early stages of training, which causes a lot of noise with the spherical harmonics.

\begin{figure*}[!t]
    \centering
    \includegraphics[width=\linewidth]{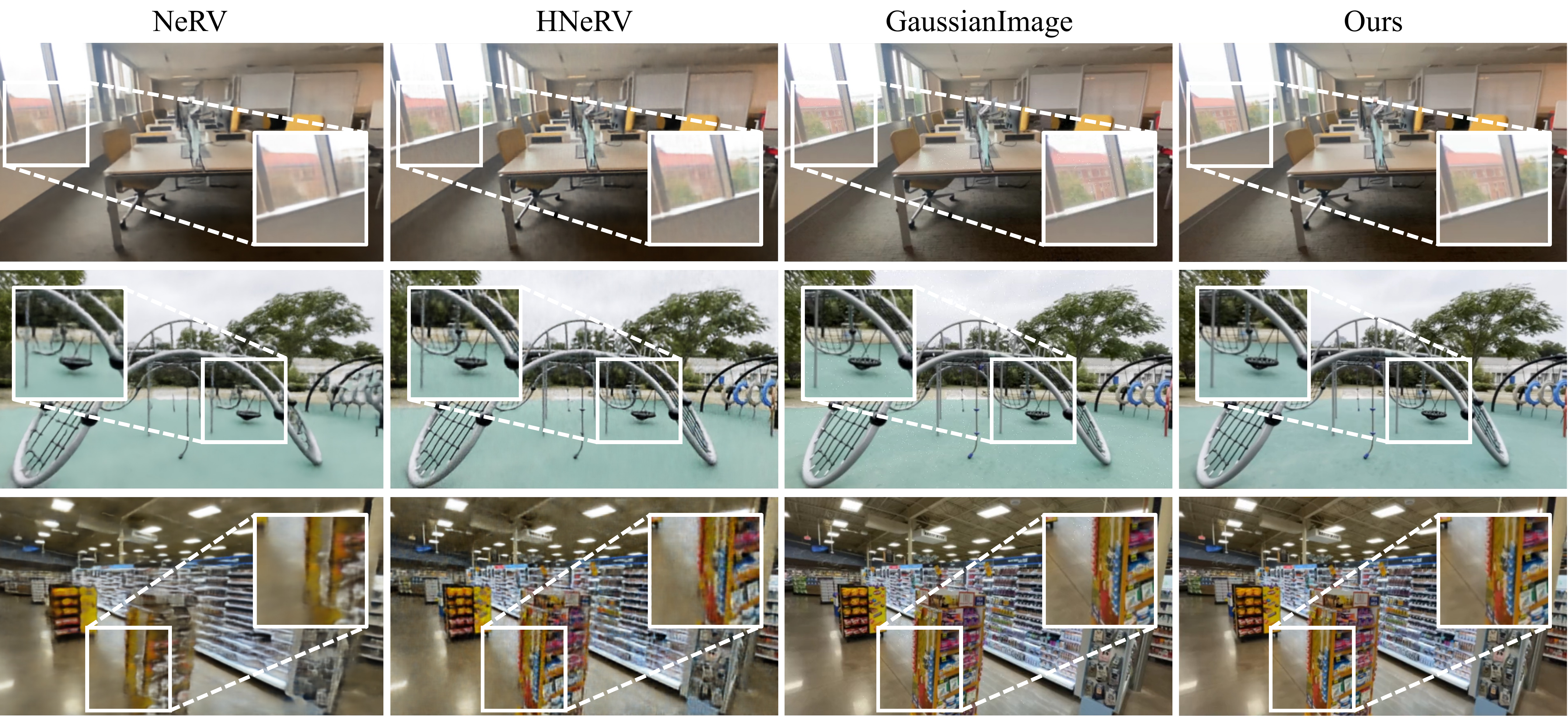}
    \caption{\textbf{Video reconstruction results on the DL3DV dataset}, comparing GaussianVideo with alternative video representation models including NeRV, HNeRV, and GaussianImage. Each sequence shows reconstructed frames, highlighting GaussianVideo's ability to capture fine spatial details and structural fidelity, even in high-motion scenes.}
    \label{fig:recon-results}
\end{figure*}

\begin{table*}[!t]
    \centering
    \begin{tabular}{lcccc}
    \midrule
     \textbf{Setting} & \textbf{PSNR} $\uparrow$ & \textbf{SSIM} $\uparrow$ & \textbf{LPIPS} $\downarrow$ & \textbf{Training Time} \\
    \midrule
    {GaussianVideo Full Model (Baseline)} & \best{43.2110} & \best{0.9882} & \best{0.0125} & $\sim$ 45 mins \\
    {Temporal Hierarchical Learning Only} & \sbest{42.8404} & \sbest{0.9871} & \sbest{0.0142} & $\sim$ 45 mins \\
    {Spatial Hierarchical Learning Only} & \tbest{42.2068} & \tbest{0.9846} & 0.0224 & $\sim$ 35 mins \\
    {No Hierarchical Learning} & 40.2016 & 0.9820 & \tbest{0.0177} & $\sim$ 40 mins\\
    {Static Camera (Fixed Learned Position)} & 33.9304 & 0.9238 & 0.1011 & $\sim$ 35 mins\\
    {No Camera Model} & 21.1821 & 0.7600 & 0.3042 & $\sim$ 20 mins\\
    {Polynomial Motion Representation} & 20.1871 & 0.6107 & 0.3991 & $\sim$ 30 mins\\
    {Incremental Spherical Harmonics Coefficients} & 31.0736 & 0.8594 & 0.1644 & $\sim$ 40 mins\\
    {Fixed Gaussian Scale Over Time} & 34.1585 & 0.9304 & 0.0951 & $\sim$ 40 mins\\
    {Static Unused Gaussians (No Warping)} & 34.0106 & 0.9247 & 0.1002 & $\sim$ 35 mins\\
    \bottomrule
    \end{tabular}%
    \caption{\textbf{Performance evaluation of various configurations in our ablation study on the DL3DV dataset}. We compare key components of the GaussianVideo pipeline such as camera representation, motion modeling, and hierarchical learning strategies. %
    }
    \label{tab:ablation}
\end{table*}

\subsection{Experimental Results}
Table \ref{tab:comparisons} compares our results with several existing methods across both datasets. Our approach consistently achieves the best scores on all three evaluation metrics for both datasets. %
Fig.~\ref{fig:recon-results} provides qualitative comparisons of reconstruction quality for the evaluated methods on three sample sequences from the DL3DV dataset. In these examples, we observe that other approaches struggle with high-frequency details, such as edges. For instance, in the store scene, other methods fail to accurately reconstruct the edges of the shelf, while our approach preserves fine details along the edge. GaussianImage, which is trained on a frame-by-frame reconstruction task, achieves good reconstruction results due to parameter scaling with the number of frames. However, this approach lacks temporal consistency across frames, as no relationship is established between Gaussians in different frames, making it difficult to maintain consistency in downstream applications.

\subsection{Analysis}
\noindent\textbf{{Ablation Study.}}
 To validate each component we introduced, we conducted an extensive ablation study across various components, summarized in Table \ref{tab:ablation}. We evaluate different configurations in our pipeline, including using a static camera instead of employing a Neural ODE, omitting camera representation entirely (as in GaussianImage), replacing splines with polynomials, applying progressive SH coefficient learning, using fixed scales over time, and avoiding the warping unrendered Gaussians. Our findings show that any form of learned camera representation is advantageous, with a substantial performance gain achieved by modeling camera motion using a Neural ODE. Additionally, we assess the impact of hierarchical learning, testing spatial-only, temporal-only, and combined hierarchical learning, as well as a setup without hierarchical learning. Both spatial and temporal hierarchical learning yield notable improvements individually, yet their combination consistently produces the best results. Notably, the spatial hierarchical learning procedure also reduces training time, as the initial phase of training involves fewer, lower-resolution Gaussians.%

\begin{figure}[!t]
    \centering
    \includegraphics[width=0.95\linewidth]{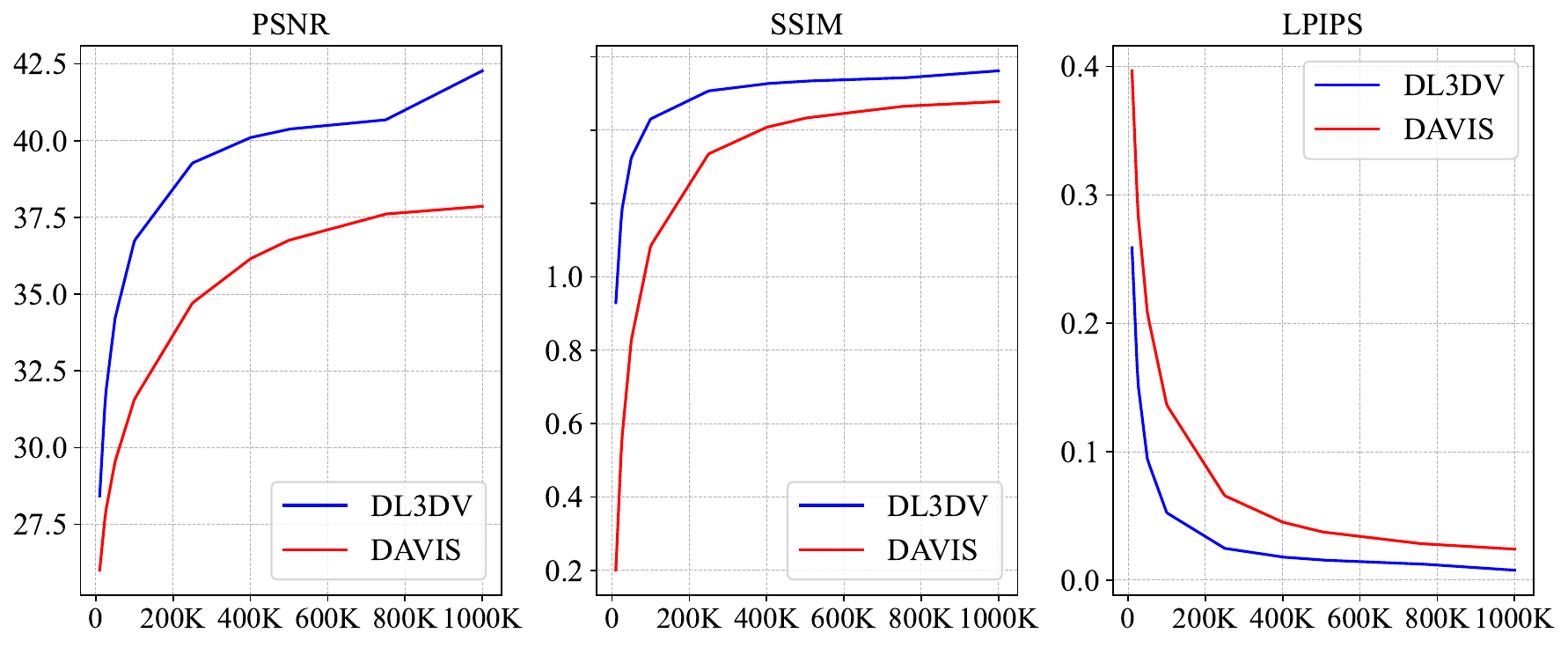}
    \caption{\textbf{Effect of Gaussian count on performance metrics.}}
    \label{fig:num_gaussians_ablation}
\end{figure}

\vspace{1mm}\noindent\textbf{Effect of Gaussian Count on Model Performance.}
To understand the influence of the number of Gaussians on performance, we test our approach on both datasets with Gaussian counts ranging from 1,000 to 1 million, training each configuration for 50K steps. As shown in Fig.~\ref{fig:num_gaussians_ablation}, even with only 100K Gaussians, our method achieves superior performance on both datasets compared to existing video-based approaches. However, performance plateaus around 400K Gaussians, beyond which further increases in Gaussian count provide minimal improvement while significantly extending training time.

\begin{figure}[!t]
    \centering
    \includegraphics[width=0.95\linewidth]{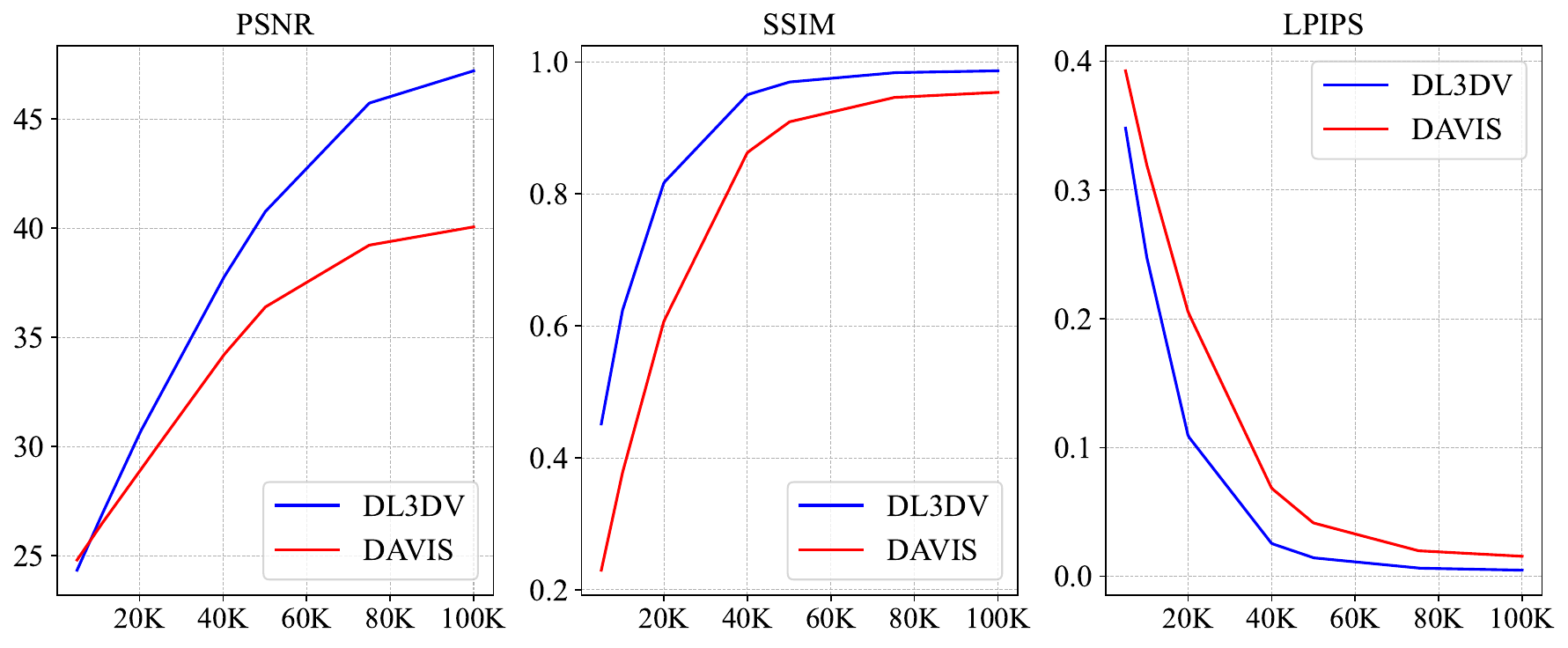}
    \caption{\textbf{Impact of training duration on performance metrics.}}
    \label{fig:training_steps_ablation}
\end{figure}

\vspace{1mm}\noindent\textbf{Impact of Training Steps on Performance Metrics.}
We also investigate the effect of training duration on performance, testing a range of training lengths from 5K to 100K steps, with all experiments utilizing 400K Gaussians. As seen in Fig.~\ref{fig:training_steps_ablation}, even with 30K timesteps, our approach outperforms the prior video-based methods. Based on these findings, we select 50K timesteps as it offers a balance between high metric performance and training efficiency.

 \begin{figure*}[!t]
    \centering
    \includegraphics[width=\linewidth]{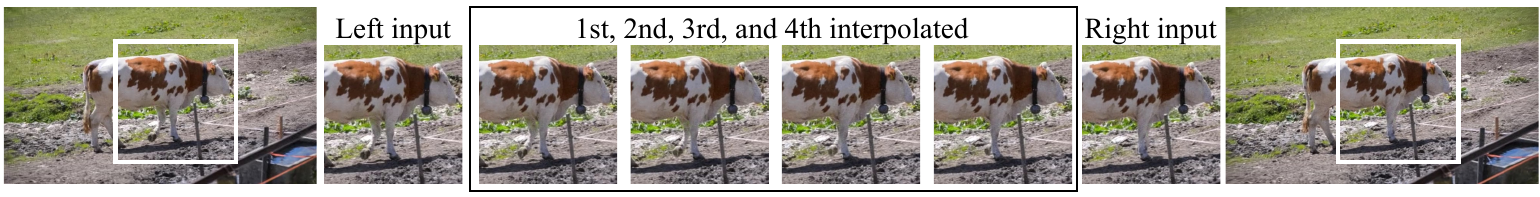}
    \caption{\textbf{Frame interpolation results for GaussianVideo using continuous motion representation.} The close-up views show the input left and right frames, along with intermediate interpolated frames, successfully capturing the realistic non-rigid motion of the cow’s legs.}
    \label{fig:interpolation}
\end{figure*}

\vspace{-2mm}\section{Applications}
\label{sec:applications}

\begin{figure}[!t]
    \centering
    \includegraphics[width=0.80\linewidth]{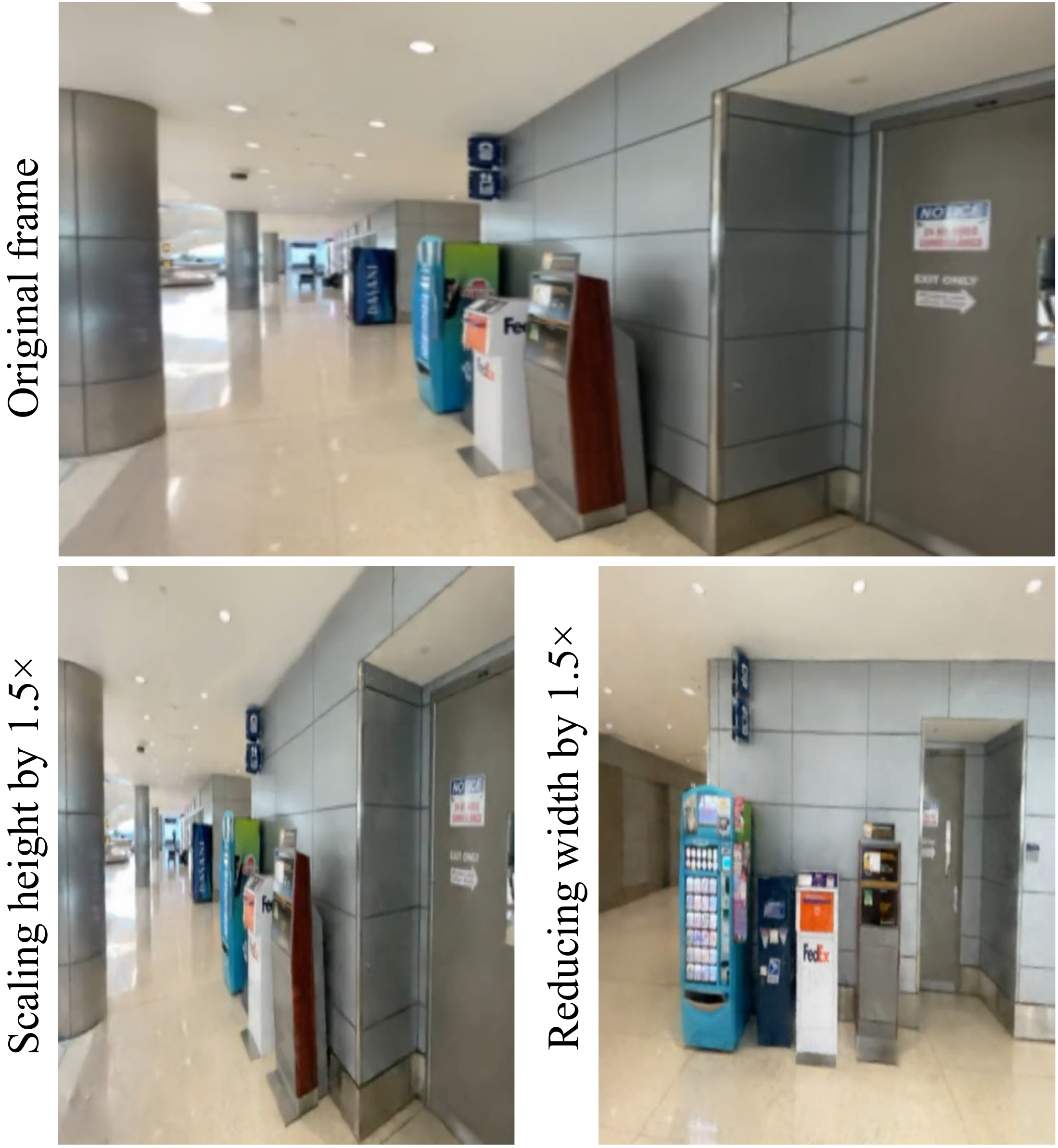}
    \caption{\textbf{Spatial resampling results on a learned video.}  In this example, the frame height is scaled by 1.5$\times$, and the width is reduced by 1.5$\times$, demonstrating the model’s ability to adjust resolution while preserving sharpness and detail. This is achieved by modifying scale parameters, as well as the principal point and focal lengths of the learned camera parameters, allowing for viewport adjustments beyond traditional resampling methods.}
    \label{fig:resampling_application}
\end{figure}

One of the advantange of our Gaussian video representation is the learned continuous nature. Although downstream applications are not the focus of the paper, we designed our method to ensure the learned Gaussian dynamics are semantic and coherent (see Fig.~\ref{fig:semantic_illustration}). To illustrate its potential, below we demonstrate a few applications.

\vspace{1mm}\noindent\textbf{Frame Interpolation.} Leveraging continuous motion representation for the Gaussians allows us to interpolate frames at arbitrary timesteps, including those not seen during training. Fig. \ref{fig:interpolation} illustrates this process, showing the input left and right frames along with four intermediate interpolated frames. These interpolated frames successfully capture the realistic non-rigid motion of the cow’s legs, demonstrating the model's ability to maintain natural motion dynamics across unseen timesteps.

\vspace{1mm}\noindent\textbf{Spatial Resampling.} The inherently spatial nature of our Gaussian video representation allows for making arbitrary adjustments in the spatial resolution of the learned videos. This is achieved by modifying the scale parameters along with the principal point and focal lengths of the learned camera parameters, using simple scaling operations. In Fig. \ref{fig:resampling_application}, we demonstrate such a spatial resampling procedure where the frame width is divided by 1.5$\times$ and the height is multiplied by 1.5$\times$. The GaussianVideo representation effectively maintains high-quality reconstruction, free from artifacts, despite the significant spatial adjustments.

\begin{figure}[!t]
    \centering
    \includegraphics[width=1.0\linewidth]{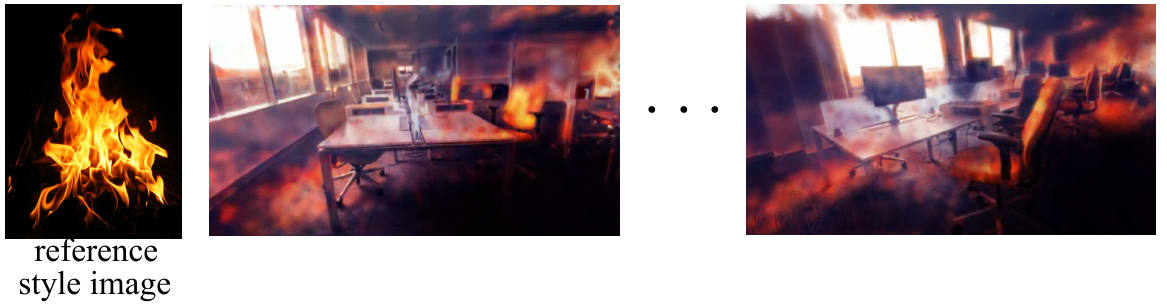}
    \caption{\textbf{Video stylization results using GaussianVideo.} Starting with a \textit{`fire'} style applied to the first frame, the style is propagated across the video by updating only the SH coefficients, ensuring consistent stylization without additional supervision. The visual coherence across frames highlights GaussianVideo’s ability to maintain high-quality style transfer while preserving structural details throughout the sequence.
    }
    \label{fig:style-transfer}
\end{figure}

\vspace{1mm}\noindent\textbf{Video Stylization.} For video stylization, we adopt the approach from \cite{splatter-a-video}. Specifically, we use an existing image editing model to edit the first frame, then propagate these changes across the entire video with a reconstruction loss, updating only the spherical harmonics coefficients. While there exist some works that perform stylization using Gaussian splatting, we focus on a straightforward approach that does not require extensive extra supervision, such as CLIP \cite{clip} losses. Fig. \ref{fig:style-transfer} shows an example where a fire image is first used to edit the initial frame, which is then propagated throughout the video in a consistent manner, such as the chair adopting a fire-like texture and the floor to the left of the table turning a vivid red.

 \vspace{-2mm}\section{Conclusion}
\label{sec:conclusion}
We introduced an efficient neural video representation method built on hierarchical Gaussian splatting, which tackles key challenges in dynamic scene representation, such as memory efficiency, reduced training times, and improved temporal consistency. By integrating B-spline-based motion trajectories with Neural ODE-driven camera modeling, our approach captures both static and dynamic elements without requiring prior camera parameters or heavy supervision. Our hierarchical learning strategy, operating across spatial and temporal domains, enables progressive refinement, yielding promising results in terms of evaluation metrics. Future work could explore the use of non-uniform rational B-spline (NURBS)~\cite{nurbs} for greater motion flexibility and further optimize the spatial hierarchy to better utilize added Gaussians.

An important practical advantage of GaussianVideo is its capability to learn camera parameters directly as part of the optimization process, which is especially valuable in single-camera settings where traditional Structure-from-Motion techniques may fall short. With its efficient parameterization, competitive training times, and high-quality reconstruction, our approach is well-suited for applications such as dynamic scene representation and video editing. Further optimization for real-time rendering could broaden its applicability to scenarios requiring high fidelity and temporal stability.

{
    \small
    \bibliographystyle{ieeenat_fullname}
    \bibliography{main}
}

\end{document}